\documentclass{article}

\usepackage{arxiv}

\usepackage[utf8]{inputenc} 
\usepackage[T1]{fontenc}    
\usepackage{hyperref}       
\usepackage{url}            
\usepackage{booktabs}       
\usepackage{amsfonts}       
\usepackage{nicefrac}       
\usepackage{microtype}      
\usepackage{lipsum}		
\usepackage{graphicx}
\usepackage{natbib}
\usepackage{doi}
\usepackage{url}
\usepackage{footnote}
\usepackage{tablefootnote} 
\makesavenoteenv{tabular}

\usepackage[usenames,dvipsnames]{xcolor}
\usepackage{array}
\usepackage{CJKutf8}
\usepackage{balance}
\usepackage{multirow}
\usepackage{subcaption}
\usepackage{placeins}
\graphicspath{ {figs/} }
\usepackage{amsmath, amssymb}
\usepackage{mathtools} 
\usepackage{autobreak}
\usepackage{type1cm}
\usepackage{environ}
\NewEnviron{myEquation}{
\begin{equation}
\scalebox{0.9}{$\BODY$}
\end{equation}
}
\usepackage{xspace}
\usepackage{bm}

\newcommand{\eg}{\textit{e.g.}}

\title{Top-down Activity Representation Learning for Video Question Answering}

\author{Yanan Wang \quad Shuichiro Haruta\quad Donghuo Zeng \quad Julio Vizcarra \quad Mori Kurokawa \vspace{0.3em} \\
{KDDI Research} \\
{{\tt \small\{wa-yanan,do-zeng,sh-haruta,xdo-zen,xju-vizcarra,mo-kurokawa\}@kddi.com} \quad
}
}

\date{}

\renewcommand{\headeright}{}

\begin{document}
\maketitle

\begin{abstract}
Capturing complex hierarchical human activities, from atomic actions (\eg, picking up one present, moving to the sofa, unwrapping the present) to contextual events (\eg, celebrating Christmas) is crucial for achieving high-performance video question answering (VideoQA).
Recent works have expanded multimodal models (\eg, CLIP, LLaVA) to process continuous video sequences, enhancing the model's temporal reasoning capabilities. However, these approaches often fail to capture contextual events that can be decomposed into multiple atomic actions non-continuously distributed over relatively long-term sequences.
In this paper, to leverage the spatial visual context representation capability of the CLIP model for obtaining non-continuous visual representations in terms of contextual events in videos, we convert long-term video sequences into a spatial image domain and finetune the multimodal model LLaVA for the VideoQA task.
Our approach achieves competitive performance on the STAR task, in particular, with a 78.4\% accuracy score, exceeding the current state-of-the-art score by \textbf{2.8} points on the NExTQA task.
\end{abstract}

\section{Introduction}
Giving a question about a video, the video question answering (VideoQA) task aims to provide an answer about the spatio-temporal visual scene. It is a promising task in achieving real-world applications including autonomous vehicles, robots, and search engines~\cite{VQA,Wang_2023_ICCV,fukui2016multimodal,ben2017mutan}.
To deeply reason about temporal visual context, one of the significant challenges is capturing complex hierarchical human activities in videos. These activities range from atomic actions (\eg, picking up a present, moving to the sofa, unwrapping the present) to contextual events (\eg, celebrating Christmas)~\cite{luo2021moma,luo2022momalrg}.

\begin{figure}[t!]
    \centering
    \includegraphics[width=0.75\columnwidth]{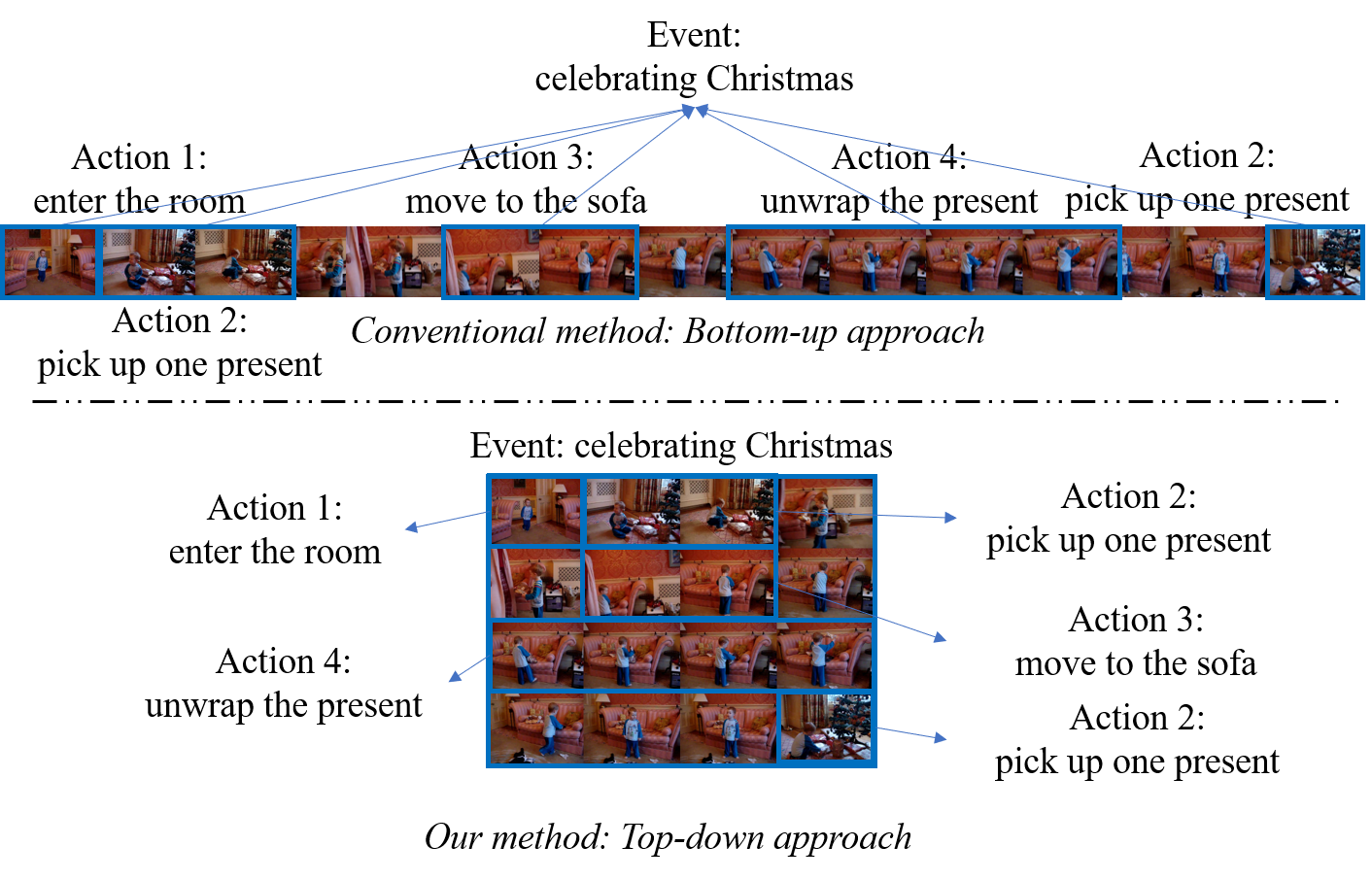}
    \caption{\textbf{Comparing different video processing approaches.} The top-down approach (ours) can leverage the strong spatial visual context representation capability to obtain effective representations of both contextual events and atomic actions.}
    \label{fig:fig1}
\end{figure}

Recently, with the dramatic advancements in large language models (LLMs) (\eg, GPT-4~\cite{achiam2023gpt}, Llama~\cite{touvron2023llama}), multimodal models such as LLaVA~\cite{liu2023llava}, an end-to-end trained large language-visual model that connects an image encoder and an LLM, have demonstrated impressive multimodal reasoning abilities on most image-based visual question answering (VQA) tasks.
To further facilitate the visual and language understanding capabilities in video scene reasoning, recent works (\eg, LLaMA-VQA~\cite{ko2023large}, Video-LLaVA~\cite{lin2023video}) have expanded multimodal models (\eg, LLaVA~\cite{liu2023llava,liu2023improvedllava}) to process continuous video sequences, enhancing the model's temporal reasoning capabilities. 
These works apply the bottom-up video processing approach to capture atomic actions in continuous short-term video sequences and lead to contextual event understanding toward complex answer reasoning.
However, these approaches often fail to capture contextual events since such events can be decomposed into multiple atomic actions that are non-continuously distributed over relatively long-term sequences (see the top of Fig.~\ref{fig:fig1}). The video sequences between adjacent atomic actions tend to be noisy, making it difficult to obtain effective contextual event representations.

This paper considers that the non-continuously distributed atomic actions can be captured by mitigating the effect of noisy video sequences in the spatial rather than the temporal domain.
Meanwhile, we can capture more effective contextual event representations with the CLIP model's powerful spatial visual context representation capability. 
To this end, we propose a top-down video processing approach that converts a long-term video sequence into a single grid image, allowing the pretrained visual encoder of the CLIP to highlight image patches representing both contextual events and atomic action pitches (see the bottom of Fig.~\ref{fig:fig1}).

We finetune the state-of-the-art multimodal model LLaVA using our proposed top-down video processing approach for VideoQA tasks.
As demonstrated by the results on two challenging benchmarks, STAR and NExTQA, our approach achieves competitive performance on the STAR task, in particular, with a 78.4\% accuracy score, exceeding the current state-of-the-art score by \textbf{2.8} points on the NExTQA task.

\section{Related work}
Instruction-following large multimodal models (LMMs) include a pretrained visual backbonesuch as CLIP~\cite{radford2021learning} to encode visual features, a pretrained large language model (LLM) such as LLaMA~\cite{touvron2023llama} to encode user instruction text and generate responses, and a projection layer used to align language and vision encoders into a common domain space. 
LLaVA~\cite{liu2023improvedllava} is a typical LMM that follows a two-stage protocol. 
In the first vision-language alignment pretraining state, LLaVA pretrains a simple MLP projection layer to align the visual encoder's output image patches with the LLM's word embedding space.
In the second visual instruction tuning stage, the model is tuned to follow the user's diverse instruction text including image content. Compared to LLaVA tuned for image-based VQA tasks, our proposed approach extends it to VideoQA tasks without adding extra training parameters.
Recent works inspired by LLaVA (\eg, Video-LLaVA~\cite{lin2023video}, Video-ChatGPT~\cite{Maaz2023VideoChatGPT}) to align video sequences with the LLM's word embedding to enable the model's temporal reasoning abilities.
Even though the short-term temporal action representation can be captured from the input of video sequences, the long-term contextual event representations are hard to obtain. Our proposed approach aims to capture the contextual event from the entire video scene.

\section{Method}
In this section, we describe our proposed top-down video processing approach and explain the training network for videoQA tasks.

\subsection{Top-down video processing}
Giving a video $V$, we sample $N^2$ frames following below sampling strategies:
\begin{enumerate}
  \item Retrieve the total number of frames $M$ in $V$ based on the video metadata such as the frame rate (fps);
  \item Split $V$ into $N^2$ intervals, and sample the middle frame from each interval;
  \item Synthesize a grid image with a size of $N\times N$ as the input of the visual encoder in the training network (see Fig.~\ref{fig:fig2}).
We synthesize grid images with sizes of $3\times 3$ for the STAR dataset and $4\times 4$ for the NExTQA dataset, respectively.
\end{enumerate}

\subsection{Training network}
\begin{figure}[th]
    \centering
    \includegraphics[width=0.7\columnwidth]{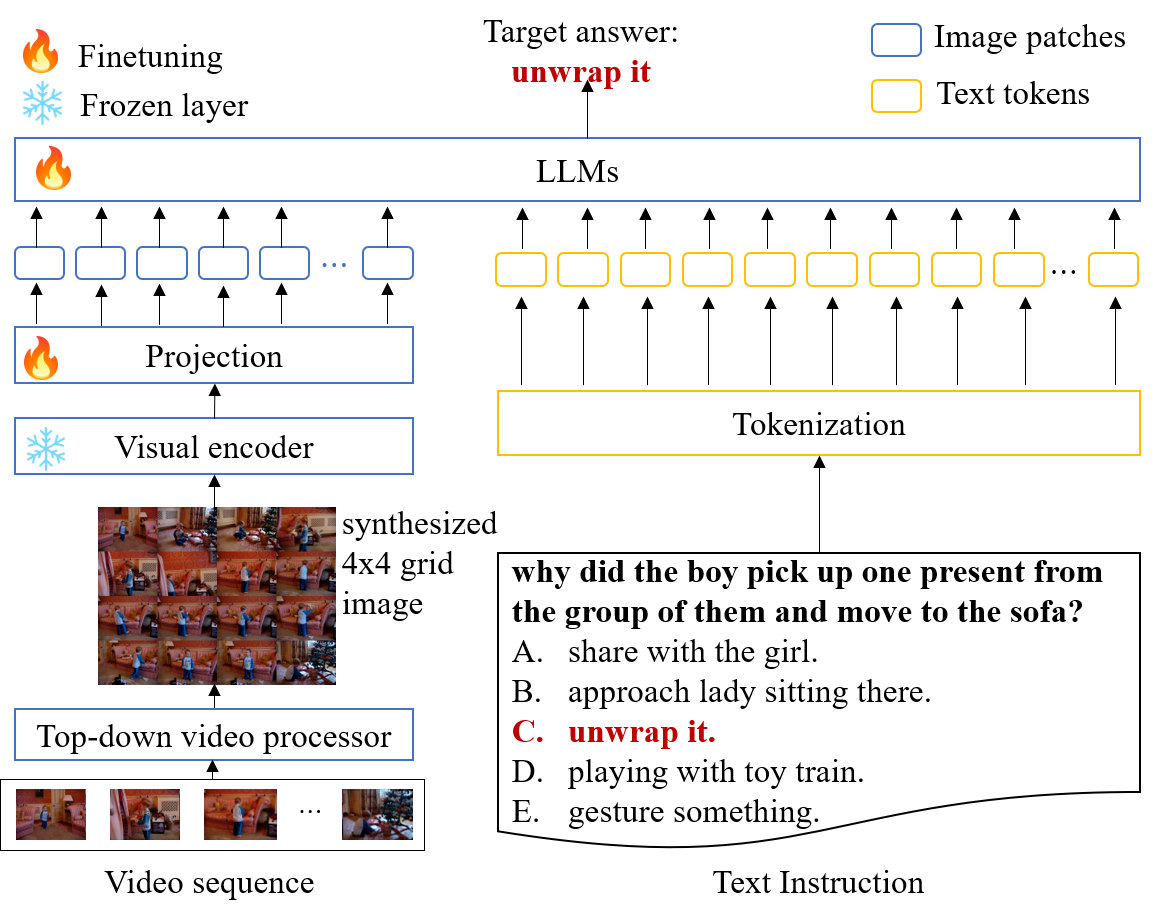}
    \caption{Overview of the training network. We add a simple top-down video processor in the LLaVA's architecture to enable the finetuning stage for achieving video question reasoning. We froze the visual encoder during the finetuning process and updated the parameters in the projection layer and LLMs.}
    \label{fig:fig2}
\end{figure}

We finetune the pretrained LLaVA model for VideoQA tasks. As shown in Fig.~\ref{fig:fig2}, we have synthesized gird images, questions, and all answer options as user instruction text. 
The input of the gird image is first resized to $336\times 336$, then embedded into $576$ image patches with the pretrained CLIP visual encoder. The pretrained CLIP works for the synthesized grid image because the image patch size is small enough to represent the fine-grained visual context.
The vision-language projector has been pretrained at the pretraining stage with image-text pairs, and the parameters are included in the pretrained LLaVA checkpoint. We finetune the pretrained projector to output the patch sequences to represent contextual representations of the entire grid image. Meanwhile, the text contents are encoded to output words' embeddings, then concatenated with image patches as the inputs of LLMs. We finetune 3 types of LLMs including Vicuna-7B~\cite{zheng2023judging}, Vicuna-13B, and Hemes-Yi-34B that have already finetuned for VQA tasks, and the pretrained parameters are also included in the pretrained LLaVA checkpoints. 
To make the model output the option's letter of correct answers directly in the inference time, we add a prompt text ``Answer with the option's letter from the given choices directly.'' into the user instruction text.
The training target is to generate the correct option. 

We utilize DeepSpeed\footnote{https://github.com/microsoft/DeepSpeed} to boost memory efficiency during the training process. We adopt the ZeRO stage 3 (ZeRO-3) to automatically collect and partition the 16-bit model parameters during the forward and backward passes. It enables us to finetune the $13B$ LLaVA model with $4\times 40GB A100$ GPUs and the $34B$ LLaVA model with $8\times 40GB A100$ GPUs.

%

\section{Experiment}
\subsection{Dataset}
We evaluate our model on two challenging VideoQA tasks including STAR~\cite{wu2021star_situated_reasoning} and NExTQA~\cite{xiao2021next}.
Both are multi-choice videoQA benchmarks featuring causal and temporal questions involving object-level spatio-temporal reasoning. NExTQA consists of causal, temporal, and descriptive question types, with a total of 5,440 videos with 47,692 questions split into training (34,132), validation (4,996), and test (8,564) sets. STAR consists of four question types for situated reasoning: Interaction, Sequence, Prediction, and Feasibility, with a total of \textbf{60K} samples split into training/validation/test sets with a ratio of about 6:1:1.

\subsection{Baselines}
\textbf{InternVideo}~\cite{wang2022internvideo} proposed a unified learning paradigm with both masked and contrastive modeling to establish a feasible and effective spatio-temporal representation.
\textbf{LLaMA-VQA}~\cite{ko2023large} proposed Flipped-VQA model to efficiently fine-tune LLMs on VideoQA by reasoning and understanding the complex relationships of ⟨V, Q, A⟩ triplet, using LLMs’ prior knowledge of temporal and causal reasoning.
\textbf{SeViLA}~\cite{yu2023self} introduced a localizer module to select top-K video frames and guide an answerer module to focus on important language-related frames and predict answers.
\textbf{ViLA}~\cite{wang2023vlap} designed a learnable text-guided Frame-Prompter together with a cross-modal distillation (QFormer-Distiller) module to select keyframes and improve the video-language alignment accuracy.
\textbf{LRR}~\cite{bhattacharyya2023look} trained an LM end-to-end on three low-level surrogate tasks, object detection, re-identification, and tracking, to endow the model with the required low-level visual capabilities.

\subsection{Results}
The main results compared with recent state-of-the-art (SoTA) scores are shown in Tab.~\ref{tab:1}. The zero-shot results produced by the original LLaVA model are worse than SoTAs since the origin LLaVA is only pretrained on image-text pairs and lacks temporal reasoning abilities. By finetuning LLaVA with our proposed top-down video processing approach, the model can tune the parameters in the visual projector and LLMs to enable learning temporal reasoning from the synthesized grid image. Our finetuning models finally achieved competitive scores for both benchmarks. In particular, on the NExTQA validation set, our model demonstrates the best score for casual and temporal question reasoning and finally achieves a new SoTA \textbf{78.4\%} accuracy score. 

\begin{table*}[h]
\centering
\resizebox{\linewidth}{!}{%
    \begin{tabular}{lcc|ccccc|cccc}
    \toprule
    \multirow{2}{*}{Model} & \multirow{2}{*}{Language model} & \multirow{2}{*}{Visual model} &  \multicolumn{4}{c}{STAR} & \multicolumn{4}{c}{NExTQA} \\ 
    \cmidrule(lr){4-12}
    &  & & Int. & Seq. & Pre. & Fea. & Tot. & Cau. & Tem. & Des. & Tot.  \\
    \midrule
    Internvideo~\cite{wang2022internvideo}
    &CLIP Text encoder-1.3B & Vanilla ViT &  62.7 & 65.5 & 54.9 & 51.9 & 58.7 & 62.5 & 58.5 & 75.8 & 63.2 \\
    LLaMA-VQA~\cite{ko2023large}
    &LLaMA-7B & CLIP ViT/L14 & 66.2& 67.9 & 57.2 & 52.7 & 65.4& 72.7 & 69.2 & 75.8 & 72.0 \\
    SeViLA~\cite{yu2023self}
    &Flan-T5 XL-3B & BLIP-4.1B & 63.7 & 70.4 & 63.1& 62.4 & 64.9 & 73.8 & 67.0 & 81.1 & 73.8 \\
    ViLA~\cite{wang2023vlap}
    & FlanT5 XL-3B&ViT-1B& 70.0 & 70.4 & 65.9 & 62.2& 67.1 & 75.3 &71.8 & 82.1 & 75.6\\
    LRR~\cite{bhattacharyya2023look}
    &OPT-0.4B& Two-stream video encoder &\textbf{73.7}&\textbf{71.0}&\textbf{71.3}&\textbf{65.1}&\textbf{70.5}&-&-&-&- \\
    \midrule
    \multicolumn{12}{c}{\textbf{Zero-shot}}  \\
    \midrule
    \multirow{2}{*}{Our approach}
    & Vicuna-13B & CLIP-ViT-L-336px & 49.4 & 49.8 & 46.3 & 42.7 & 48.9 & 64.8 & 62.3 & 68.2 & 64.5\\
    & Hermes-Yi-34B & CLIP-ViT-L-336px & 49.5 & 50.3 & 47.0 & 41.6 & 49.1 & 71.7 & 66.8 & 75.2 & 70.6\\
    \midrule
    \multicolumn{12}{c}{\textbf{Finetuning}} \\
    \midrule
    \multirow{2}{*}{Our approach}
    & Vicuna-13B & CLIP-ViT-L-336px & 66.9 & 68.3 & 59.1 & 56.3 & 66.2 & 75.7 & 70.8 & \textbf{82.1} & 75.1\\
    & Hermes-Yi-34B & CLIP-ViT-L-336px & 66.2 & 69.6 & 60.6 & 55.5 & 66.7 & \textbf{80.1} & \textbf{75.6} & 78.6 & \textbf{78.4}\\
    \bottomrule

    \end{tabular}
}
\captionsetup{width=\textwidth}\caption{Comparison accuracy results (\%) with SoTAs on the validation set of STAR and NExT-QA datasets. Based on our proposed top-down video processing approach, the finetuned LLaVA model for the NExT-QA dataset achieves the new SoTA that is \textbf{2.8\%} higher accuracy than the ViLA model. Here, Int., Seq., Pre., Fea., Cau., Tem., and Des. are abbreviations of  Interaction, Sequence, Prediction, Feasibility, Causal, Temporal, and Descriptive question types, respectively. Total is abbreviated as Tot..  *The 34B model for NExTQA is finetuned with the LoRA training strategy.}
\label{tab:1}
\end{table*}

\subsection{Case study}
\begin{figure*}[t]
    \centering
    \includegraphics[width=\textwidth]{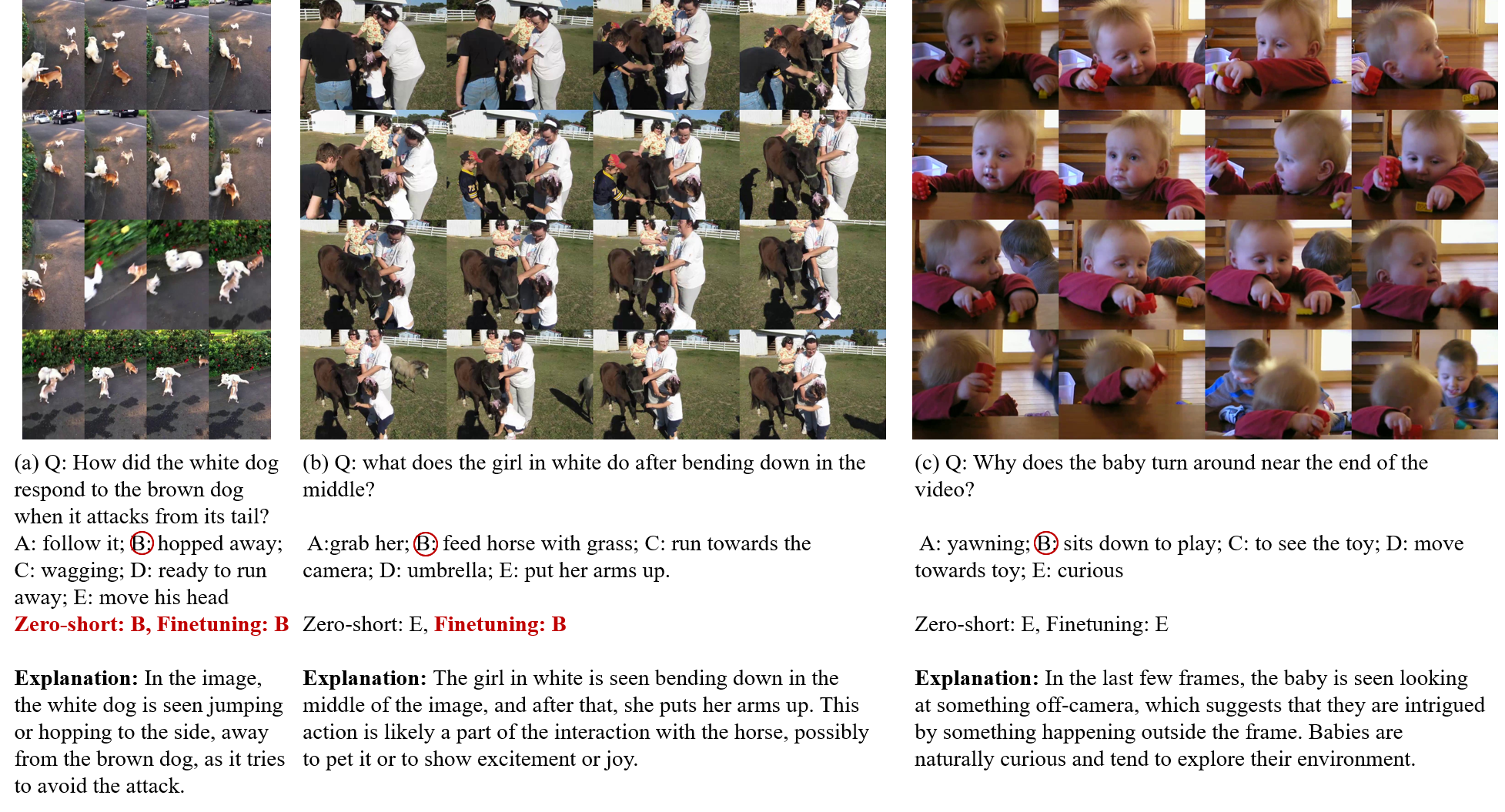}
    \captionsetup{width=\textwidth}\caption{Three cases selected from NExTQA validation set demonstrate the effect of our proposed video processing approach on video scene understanding. Here, we synthesize $4\times 4$ grid images and use them as the input. The explanation comes from the zero-shot model.}
    \label{fig:fig3}
\end{figure*}

To further investigate the effect of our proposed top-down video processing approach on video scene understanding, we compared zero-shot and finetuning models with three data samples in Fig.~\ref{fig:fig3}. 
We noticed that the zero-shot model works well when the context of different video frames changes significantly. As a result, the model can easily capture temporal actions from changes in context (see Fig.~\ref{fig:fig3}(a)).
In contrast, the zero-shot model fails the sample in Fig.~\ref{fig:fig3}(b), since the model can not reason the high-level event without tuning the model parameters on the synthesized grid images. For example, the zero-shot model can only capture a low-level action ``put her arms up'', but fails to reason a high-level event ``feed horse with grass''. On the other hand, as the finetuning model learned the holistic context in the video from the given grid images, it achieved the correct answer.
For the sample in Fig.~\ref{fig:fig3}(c), both zero-shot and finetuning models fail the question. The reason is that the relevant frames to the target action are not picked up correctly from the actual video. This highlights the importance of extracting text-related context from a long-term video.
We also prompt the zero-shot model to explain the answer in Fig.~\ref{fig:fig3}. By leveraging the powerful pretrained LLaVA model, the zero-shot model can provide a reasonable explanation. However, we noticed that the finetuning model fails to output an explanation since it overfits the target VideoQA task and forgets the pretrained knowledges. We plan to focus on this crucial issue in the next step. 

\subsection{Ablation study}
To further study how the synthesized grid image is effective in drawing temporal reasoning of LLaVA out, we compared different sizes of the synthesized grid image in Tab.~\ref{fig:fig2}. The result demonstrates the $3\times 3$ grid image including 9 frames performs better than others. In particular, when we randomly select one frame, the result gets worse since the pretrained LLM can not perform temporal reasoning with a single frame. In contrast, the processed grid image enables the model to capture the top-down visual context in the video scene. 

We also compared our proposed top-down video processing approach with its bottom-up counterpart in Tab.~\ref{tab:3}.
Here, top-down video processing aims to capture the context of the entire video, and its bottom-up counterpart focuses on aggregating the frame-level context of the video sequence. By comparing two approaches built on LLaMA-based LLM with the same model size, our approach achieves \textbf{82.1\%} accuracy for the description question and exceeds \textbf{6.2\%} of its bottom-up counterpart. This result demonstrates the efficacy of our approach for the holistic understanding of the video scene.

\begin{table}[h]
\centering
    \begin{tabular}{l|ccccc}
    \toprule
    Model &  Int. & Seq. & Pre. & Fea. & Tot.  \\
    \midrule
    Zero-shot(13B)-1-frame & 46.8 & 43.6 & 38.5 & \textbf{43.1} & 44.2 \\
    Zero-shot(13B)-16-frames & 47.8 & 45.9 & 45.2 & 41.4 & 46.2 \\
    Zero-shot(13B)-9-frames & \textbf{49.4} & \textbf{49.8} & \textbf{46.3} & 42.7 & \textbf{48.9} \\
    \bottomrule
    \end{tabular}
\caption{The zero-shot evaluation results on the STAR dataset demonstrate the efficacy of the proposed top-down video processing. Here, ``Zero-shot(13B)-\textbf{\{xx\}}-frames'' denotes the number of video frames included in the synthesized grid image.}
\label{tab:2}
\end{table}

\begin{table}[h]
\centering
    \begin{tabular}{l|ccccc}
    \toprule
    Model & Cau. & Tem. & Des. & Tot.  \\
    \midrule
    Bottom-up-13B-LLaMA &75.3 & \textbf{71.7} &75.9 &74.2 \\
    Top-down-13B-LLaMA & \textbf{75.7} & 70.8 & \textbf{82.1} & \textbf{75.1} \\
    \bottomrule
    \end{tabular}
\caption{Comparison of different video processing strategies.}
\label{tab:3}
\end{table}

\subsection{Limitation}
Even though recent multimodal models demonstrate powerful visual reasoning abilities, they fail to detect objects correctly and result in incorrect video context reasoning. In addition, the finetuned model tends to forget pretrained knowledge, which can worsen the robustness of the model in real-world applications.

\section{Conclusion}

In this paper, we proposed a top-down video processing approach that converts a long-term video sequence into a single grid image, allowing the pretrained visual encoder of the CLIP to highlight image patches representing both contextual events and atomic action pitches. By finetuning a multimodal model LLaVA on the synthesized grid images, we achieved competitive performance on the STAR task, in particular, with a 78.4\% accuracy score, exceeding the current state-of-the-art score by \textbf{2.8} points on the NExTQA task. In future work, we are working on how to deal with how to capture low-level video representations to boost object recognition and tracking tasks.

\bibliographystyle{unsrtnat}
\bibliography{references}  






\end{document}